\documentclass{article}
\usepackage{ijcai19}

\usepackage{times}
\usepackage{soul}
\usepackage{url}
\usepackage[utf8]{inputenc}
\usepackage[small]{caption}
\usepackage{graphicx}
\usepackage{amsmath}
\usepackage{booktabs}
\usepackage[usenames,dvipsnames,svgnames,table]{xcolor}
\DeclareGraphicsExtensions{.png}
\usepackage{latexsym}
\usepackage{xspace}
\usepackage{listings}
\usepackage{enumitem}
\usepackage{balance}
\usepackage{algorithm}
\usepackage{algorithmic}
\usepackage{mathtools}
\usepackage{amssymb}
\usepackage[position=bottom]{subfig}
\usepackage{float}
\usepackage{mathabx}
\usepackage{array,multirow}

\newtheorem{example}{Example}

\newcommand*{\eat}[1]{}

\newcommand*{\agg}{\textsf{IGF-Aggregated}\xspace}
\newcommand*{\ratio}{\textsf{IGF-Ratio}\xspace}

\usepackage{color}

\newcommand{\constr}{diversity constraints\xspace}

\newcommand{\igf}{in-group fairness\xspace}

\title{Balanced Ranking with Diversity Constraints}
\author{
Ke Yang$^1$\footnote{Contact Author}\and
Vasilis Gkatzelis$^2$\and
Julia Stoyanovich$^1$\\
\affiliations
$^1$New York University, Department of Computer Science and Engineering\\
$^2$Drexel University, Department of Computer Science\\
\emails ky630@nyu.edu, gkatz@drexel.edu, stoyanovich@nyu.edu}

\begin{document}

\maketitle
\begin{abstract}
Many set selection and ranking algorithms have recently been enhanced with {\em \constr} that aim to explicitly increase representation of historically disadvantaged populations, or to improve the overall representativeness of the selected set. An unintended consequence of these constraints, however, is reduced {\em \igf}: the selected candidates from a given group may not be the best ones, and this unfairness may not be well-balanced across groups.
In this paper we study this phenomenon using datasets that comprise multiple sensitive attributes. We then introduce additional constraints, aimed at balancing the \igf across groups, and formalize the induced optimization problems as integer linear programs. Using these programs, we conduct an experimental evaluation with real datasets, and quantify the feasible trade-offs between balance and overall performance in the presence of \constr. 
\end{abstract}
\section{Introduction}
\label{sec:intro}

The desire for diversity and fairness in many contexts, ranging from results of a Web search to admissions at a university, has recently introduced the need to revisit algorithm design in these settings. Prominent examples include set selection and ranking algorithms, which have recently been enhanced with \textit{\constr} 
\cite{DBLP:conf/icalp/CelisSV18,drosou2017diversity,stoyanovich2018online,DBLP:journals/corr/ZehlikeB0HMB17,DBLP:journals/datamine/Zliobaite17}. 
Such constraints focus on groups of items in the input that satisfy a given \textit{sensitive attribute label}, typically denoting membership in a demographic group, and seek to ensure that these groups are appropriately represented in the selected set or ranking.  Notably, each item will often be associated with multiple attribute labels (e.g., an individual may be both female and Asian).

Diversity constraints may be imposed for legal reasons, such as for compliance with Title VII of the Civil Rights Act of 1964~\cite{titlevii}.  Beyond legal requirements, benefits of diversity, both to small groups and to society as a whole, are increasingly recognized by sociologists and political scientists  ~\cite{diff,crowdsourcing}.  Last but not least, diversity constraints can be used to ensure dataset representativeness, for example when selecting a group of patients to study the effectiveness of a medical treatment, or to understand the patterns of use of medical services~\cite{cohen2009medical}, an example we will revisit in this paper.

Our goal in this paper is to evaluate and mitigate an unintended consequence that such \constr may have on the outcomes of set selection and ranking algorithms. Namely, we want to ensure that these algorithms do not systematically select lower-quality items in particular groups.  In what follows, we make our set-up more precise.

Given a set of items, each associated with multiple sensitive attribute labels and with a quality score (or utility), a {set selection} algorithm needs to select $k$ of these items aiming to maximize the overall utility, computed as the sum of \textit{utility scores} of  selected items. The score of an item is a single scalar that may be pre-computed and stored as a physical attribute, or it may be computed on the fly. The output of traditional set selection algorithms, however, may lead to an underrepresentation of items with a specific sensitive attribute. As a result, recent work has aimed to modify these algorithms with the goal of introducing diversity.

There are numerous ways to define diversity. For set selection, a unifying formulation for a rich class of proportional representation fairness~\cite{DBLP:journals/datamine/Zliobaite17} and coverage-based diversity~\cite{drosou2017diversity} measures is to specify a lower bound $\ell_v$ for each sensitive attribute value $v$, and to enforce it as the minimum cardinality of items satisfying $v$ in the output \cite{stoyanovich2018online}. 
If the $k$ selected candidates need to also be ranked in the output, this formulation can be extended to specify a lower bound  $\ell_{v,p}$ for every attribute $v$ and every prefix $p$ of the returned ranked list, with $p \leq k$~\cite{DBLP:conf/icalp/CelisSV18}. Then, at least $\ell_{v,p}$ items satisfying $v$ should appear in the top $p$ positions of the output. We refer to all of these as {\em \constr} in the remainder of this paper. 
Given a set of \constr, one can then seek to maximize the utility of the selected set, subject to these constraints.

As expected, enforcing \constr often comes at a price in terms of overall utility.
Furthermore, the following simple example exhibits that, when faced with a combination of \constr,
maximizing utility subject to these constraints can lead to another form of imbalance.

\begin{table}[]
\centering
\begin{tabular}{|l|c|c|c|c|}
   \hline
   \rowcolor{LightGrey}
     & \multicolumn{2}{c|}{\bf Male} &  \multicolumn{2}{c|}{\bf Female}  \\
   \hline
   White &  A (99) &  B (98) &  C (96) & D (95) \\
   Black &  E (91) &  F (91) &  G (90) & H (89) \\
   Asian &  I (87) &  J (87) &  K (86) & L (83) \\
   \hline
\end{tabular}
\caption{A set of 12 individuals with sensitive attributes {\tt race} and {\tt gender}.  
Each cell lists an individual's ID, and score in parentheses.}
\label{tab:example_constraints}
\end{table}
\begin{example}
Consider $12$ candidates who are applying for $k=4$ committee positions. Table~\ref{tab:example_constraints} illustrates this example using a letter from A to L as the candidate ID and specifying the ethnicity, gender, and score of each candidate (e.g., candidate E is a Black male with a score of 91). 
Suppose that the following \constr are imposed: the committee should include two male and two female candidates, and at least one candidate from each race. In this example both race and gender are strongly correlated with the score: White candidates have the highest scores, followed by Black and then by Asian. Further, male candidates of each race have higher scores than female candidates of the same race.

The committee that maximizes utility while satisfying the \constr is $\{\text{A, B, G, K}\}$, with utility score $373$. Note that this outcome fails to select the highest-scoring female candidates (C and D), as well as the highest-scoring Black (E and F) and Asian (I and J) candidates. This is in contrast to the fact that it selects the best male and the best White candidates (A and B). This type of ``unfairness'' is unavoidable due to the \constr, but in this outcome it hurts historically disadvantaged groups (e.g., females and Black candidates) more.
However, one can still try to distribute this unfairness in a more ``balanced'' way across different sensitive attribute values. For instance, an alternative committee selection could be $\{\text{A, C, E, K}\}$, with utility 372. For only a small drop in utility, this ensures that the top female, male, White, and Black candidates are selected.
\label{ex:1}
\end{example}

Example~\ref{ex:1} illustrates that \constr may inevitably lead to unfairness within groups of
candidates. An important concern is that, unless appropriately managed, this unfairness may disproportionately affect 
demographic groups with lower scores. This is particularly problematic when lower scores are the 
result of historical disadvantage, as may, for example, be the case in standardized testing~\cite{doi:10.1177/0160597615603750}.
In this paper, we focus on this phenomenon and our goal is to provide measures for quantifying fairness in this context, to which we refer as \emph{\igf}, and to study the extent to which its impact can be \emph{balanced} across groups rather than disproportionately affect a few groups.

\paragraph*{Contributions} We make the following contributions:
\begin{itemize}
\item We observe that \constr can impact \igf, and introduce two novel measures that quantify this impact. 
\item  We observe that the extent to which \igf is violated in datasets with multiple sensitive 
attributes may be quite different for different attribute values.
\item We translate each of our \igf measures into a set of constraints for an integer linear program. 
\end{itemize}

{\bf Organization.}
After providing the required definitions and notation in Section~\ref{sec:preliminaries}, we introduce our notions of \igf in Section~\ref{sec:constr}.
We translate \igf into constraints for an integer linear program in Section~\ref{sec:minimization}, and use the leximin criterion to select 
the best feasible parameters. We present experiments on real datsets in Section~\ref{sec:experiment}, discuss related work in Section~\ref{sec:related}, and conclude in Section~\ref{sec:conc}.
\section{Preliminaries and Notation}
\label{sec:preliminaries}
Both the set selection and the ranking problem are defined given a set $I$ of $n$ \textit{items} (or candidates), along with a score $s_i$ associated with each item $i\in I$; this score summarizes the qualifications or the relevance of each item. The goal of the \emph{set selection} problem is to choose a subset $A\subseteq I$ of $k\ll n$ of these items aiming to maximize the total score $\sum_{i\in A}s_i$. The \emph{ranking} problem, apart from selecting the items, also requires that the items are ranked --- assigned to distinct positions $1,2,\dots,k$. In this paper we study the impact of \constr that may be enforced on the outcome.

Each item is labeled based on a set $Y$ of sensitive attributes. For instance, if the items correspond to job candidates, the sensitive attributes could be ``race'', ``gender'', or ``nationality''. Each attribute $y \in Y$ may take one of a predefined set (or domain) of values, or labels, $L_y$; for the attribute ``gender'', the set of values would include ``male'' and ``female''. We refer to attributes that have only two values (e.g., the values $\{\text{male, female}\}$ for ``gender'') as \emph{binary}. We use $L =\bigcup_{y\in Y} L_y$ to denote the set of all the attribute values related to attributes in the set $Y$. (To simplify notation, we assume that domains of attribute values do not overlap.)

Given a sensitive attribute value $v\in L$, we let $I_v \subseteq I$ denote the set of items that satisfy this label. For instance, if $v$ corresponds to the label ``female'', then $I_v$ is the set of all female candidates. We refer to such a set $I_v$ as a \emph{group} (e.g., the group of female candidates). 
For each attribute value $v$ and item $i\in I_v$, we let $I_{i,v}=\{j\in I_v:~ s_j\geq s_i\}$ be the set of items with attribute value $v$ that have a score greater than or equal to $s_i$ (including $i$); for simplicity, and without loss of generality, we assume that no two scores are exactly equal. 
In Example~\ref{ex:1} if we let $v$ correspond to the attribute value ``Black'' and let $i$ be candidate G, then $I_{i,v}=\{\text{E, F, G}\}$ and $S_{i,v}=272$.
We also use $s_{\max}=\max_{i\in I}\{s_i\}$ and $s_{\min}=\min_{i\in I}\{s_i\}$ to denote the maximum and minimum scores over all available items. Let $\lambda =s_{\max}/s_{\min}$ be the ratio of the maximum over the minimum score. 

For a given selection, $A$, of $k$ items, we use $A_v \subseteq I_v$ to denote the subset of items in $I_v$ that are in $A$, and $B_v\subseteq I_v$ for those that are not. Also, $a_v =\min_{i\in A_v}\{s_i\}$ is the lowest score among the ones that were accepted and $b_v =\max_{i\in B_v}\{s_i\}$ the highest score among the items in $I_v$ that were rejected. We say that a set selection of $k$ items is \emph{in-group fair} with respect to $v$ if $b_v \leq a_v$, i.e., no rejected candidate in $I_v$ is better than an accepted one. In Section~\ref{sec:constr} we define two measures for quantifying how in-group fair a solution is. Finally, we use $A_{i,v}=A_v \cap I_{i,v}$ to denote the set of selected items in $I_v$ whose score is at least $s_i$ for some $i\in A_v$.

\section{Diversity and In-Group Fairness}
\label{sec:constr}
To ensure proportional representation of groups in algorithmic outcomes, one way of introducing \constr is to specify lower bounds for each group $I_v$. 
Using \constr, one may require, for example, that the proportion of females in a selected set or in the top-$p$ positions of a ranking, for some $p\leq k$, closely approximates the proportion of females in the overall population. Depending on the notion of appropriate representation, one can then define, for each sensitive attribute value (or label) $v$ and each position $p$, a lower bound $\ell_{v,p}$ on the number of items satisfying value $v$ that appear in the top-$p$ positions of the final ranking (see, e.g.,~\cite{DBLP:conf/icalp/CelisSV18}). Given such bounds, one can then seek to maximize the overall utility of the generated set or ranking. The main drawback of this approach is that it may come at the expense of \igf, described next.

Given some label $v$ and the group of items $I_v$ that satisfy this label, \igf requires that if $k_v$ items from $I_v$ appear in the outcome, then it should be the \emph{best} $k_v$ items of $I_v$. Note that, were it not for \constr, any utility maximizing outcome would achieve \igf: excluding some candidate in favor of someone with a lower score would clearly lead to a sub-optimal outcome. However, as we verified in Example~\ref{ex:1}, in the presence of \constr, maintaining \igf for all groups may be impossible. In this paper we propose and study two measures of approximate \igf, which we now define.
\paragraph*{\ratio} 
Given an outcome $A$, a direct and intuitive way to quantify \igf for some group $I_v$ is to consider the unfairness 
in the treatment of the most qualified candidate in $I_v$ that was non included in $A$. Specifically, we use the ratio between the minimum score
of an item in $A_v$ (the lowest accepted score) over the maximum score in $B_v$ (the highest rejected score). This provides a fairness measure with values in the range $[0, 1]$ and 
higher values implying more \igf. For instance, if some highly qualified candidate in $I_v$ with score $s$ was rejected in favor of another candidate in $I_v$ with score $s/2$, 
then the ratio is $1/2$.

\begin{equation}
\ratio(v) = \frac{a_v}{b_v}
\label{eq:ratio_q}
\end{equation}

\paragraph*{\agg} 
Our second measure of \igf aims to ensure that for every selected item $i\in A_v$ the score $\sum_{h\in A_{i,v}} s_h$ is a good approximation of $\sum_{h\in I_{i,v}} s_h$. The first sum corresponds to aggregate utility of all accepted items $h\in I_v$ with $s_h\geq s_i$, and the second sum is the aggregate utility of all items (both accepted and rejected) $h\in I_v$  with $s_h\geq s_i$. If no qualified candidate in $I_v$ is rejected in favor of some less qualified one, then these sums are equal. On the other hand, the larger the fraction of high-scoring candidates rejected in favor of lower-scoring ones, the wider the gap between these sums. For a given group $I_v$, our second \igf measure is the worst-case ratio (over all $i\in A_v$) of the former sum divided by the latter. Just like our first measure, this leads to a number in the range $[0,1]$, with greater values indicating more \igf.
\begin{equation}
\agg(v) = \min_{i\in A_v}\left\{ \frac{\sum_{h\in A_{i,v}}s_h}{\sum_{h\in I_{i,v}}s_h} \right\}
\label{eq:agg_q}
\end{equation}
\section{Balancing In-Group Fairness}
\label{sec:minimization}
As our experiments in Section~\ref{sec:experiment} show, the distribution of \igf across groups can be quite imbalanced in the presence
of diversity constraints. In order to mitigate this issue, we now introduce
additional constraints that aim to better distribute \igf across groups. We begin by showing that the problem of 
maximizing the total utility subject to a combination of diversity and \igf constraints can be formulated as 
an integer linear program. 

It is worth noting that even the much simpler problem of checking the feasibility of a given set of diversity constraints
(let alone maximizing total utility and introducing additional \igf constraints) is NP-hard (see Theorem 3.5 in \cite{DBLP:conf/icalp/CelisSV18}). Therefore, we cannot not expect to
solve our optimization problems in polynomial time. In light of this intractability, we instead formulate our optimization
problems as integer programs. Although solving these programs can require exponential time in the worst case, standard integer programming libraries allow us to solve reasonably large instances in a very small amount of time. We briefly discuss the computational demands of these programs in Sec.~\ref{sec:experiment}. 

\subsection{Integer Program Formulations}
\label{sec:ilp}

The integer programs receive as input, for each $v\in L$ and $p\in [k]$, a set of $\ell_{v,p}$ values, and for each $v\in L$,
a $q_v$ value, which correspond to diversity and \igf lower bounds, respectively. The output of the program is a
utility-maximizing ranking of $k$ items such that at least $\ell_{v,p}$ items from $I_v$ are in the top $p$ positions, 
and the \igf of each $v\in L$ is at least $q_v$. For each item $i\in I$ and each position $p\in [k]$ of the ranking, 
the integer program uses an indicator variable $x_{i,p}$ that is set either to 1, indicating that item $i$ is in position 
$p$ of the ranked output, or to 0, indicating that it is not. We also use variable $x_i = \sum_{p\in [k]} x_{i,p}$
that is set to 1 if $i$ is included in any position of the output, and to 0 otherwise. The program below is for the \ratio measure.

\begin{equation*}
\begin{split}
\text{maximize:}  \ \sum_{i\in I} x_{i}s_i ~~~~~~~~~~~~~~~~~~ \\
\text{subject to:} ~x_i = \sum_{p \in [k]} x_{i,p}, \ ~~~~ \forall i\in I \\
				\sum_{i\in I} x_{i,p} \leq 1, \ ~~~~ \forall p\in [k] \\
        \sum_{i\in I_v} \sum_{q \in [p]} x_{i,q}\geq \ell_{v,p}, \ ~~~~ \forall v\in L, \forall p\in [k] \\
        a_v\leq (\lambda - (\lambda-1) \cdot x_i) s_i, \ ~~~~ \forall v\in L, \forall i\in I_v\\
        b_v\geq (1 - x_i) s_i, \ ~~~~ \forall v\in L, \forall i\in I_v\\
				a_v \geq q_v b_v, \ ~~~~~~~~~~~~~~~~~ \forall v\in L \\
        a_v, b_v \in [s_{\min}, s_{\max}],\ ~~~~~~~~~~~~~~~~~ \forall v\in L\\
        x_i, x_{i,p}\in \{0,1\},\ ~~~~ \forall i\in I, \forall p\in [k]\\
\end{split}
\end{equation*}

The first set of inequalities ensures that at most one item is selected for each position. The second set of inequalities corresponds to the \constr: for each attribute value $v$ and position $p$, include at least $\ell_{v,p}$ items from $I_v$ among the top $p$ positions of the output. Most of the remaining constraints then aim to guarantee that every group $I_v$ has an \ratio of at least $q_v$. Among these constraints, the most interesting one is $a_v\leq (\lambda - (\lambda-1) \cdot x_i) s_i$. This is a linear constraint, since both $\lambda= s_{\max}/s_{\min}$ and $s_i$ are constants. Note that, if $x_i$ is set to be equal to 1, i.e., if item $i\in I$ is accepted, then this constraint becomes $a_v\leq s_i$ and hence ensures that $a_v$ can be no more than the score of any accepted item. Yet, if $x_i$ is set to be equal to 0, i.e., if item $i$ is rejected, then this becomes $a_v\leq \lambda s_i = s_{\max} \frac{s_i}{s_{\min}} \leq s_{\max}$, and hence this constraint does not restrict the possible values of $a_v$. Therefore, $a_v$ is guaranteed to be no more than the smallest accepted score in $I_v$ and $b_v$ is at least as large as the smallest rejected score, since $b_v\geq (1 - x_i) s_i$, and hence $b_v \geq s_i$ if $x_i=0$. As a result, $a_v \geq q_v b_v$ enforces the \ratio constraint $\frac{a_v}{b_v}\geq q_v$.

In order to get the integer linear program that enforces \agg constraints, we modify the program above by replacing all the constraints that involve $a_v$ or $b_v$ with the set of constraints $\sum_{h\in I_{iv}} x_{h}s_h \geq q_v  x_{i}  \sum_{h\in I_{iv}} s_h$ for all $v\in L$ and all $i\in I_v$. Note that for $i$ with $x_i=0$ this constraint becomes trivially satisfied, since the right hand side becomes 0 and the left hand side is always positive. If, on the other hand, $x_i=1$, then this inequality ensures that, for all $i\in A_v$:
\[\frac{\sum_{h\in I_{i,v}} x_{h}s_h}{ \sum_{h\in I_{i,v}} s_h} ~\geq~ q_v ~~\Rightarrow~~  \frac{\sum_{h\in A_{i,v}} s_h}{ \sum_{h\in I_{i,v}} s_h} ~\geq~ q_v.\]

\subsection{The Leximin Solution}
\label{sec:leximin}

Equipped with the ability to compute the utility-maximizing outcome
in the presence of both \igf and \constr, our next step is to use
these integer linear programs to optimize the balance of \igf across 
different groups. In fact, this gives rise to a non-trivial instance of a 
fair division problem: given one of the \igf measures that we have defined, 
for each possible outcome, this measure quantifies how ``happy'' each group 
should be, implying a form of ``group happiness''. Therefore, each outcome yields a vector $\mathbf{q}=(q_1, q_2, \dots, q_{|L|})$ of \igf values for each group. As we have seen, achieving an \igf vector of $\mathbf{q}=(1, 1, \dots, 1)$ may be infeasible due to the diversity constraints. Given the set $Q$ of feasible vectors $\mathbf{q}$ implied by the diversity constraints, our goal is to identify an outcome (i.e., an \igf vector $\mathbf{q}$) that fairly distributes \igf values across all groups. 

From this perspective, our problem can be thought of as a fair division problem where the goods being allocated are the $k$ slots in the set, and the fairness of the outcome towards all possible groups is evaluated based on the \igf vector $\mathbf{q}$ that it induces. Fair division is receiving significant attention in economics~(e.g.,~\cite{Moulin03}) and
recently also in computer science~(e.g.,~\cite[Part II]{Comsoc}). A common solution that this
literature provides to the question of how happiness should be distributed is \emph{maximin}.
Over all feasible \igf vectors $\mathbf{q}\in Q$, the maximin solution dictates that a fair solution should
maximize the minimum happiness, so it outputs $\arg\max_{\mathbf{q}\in Q}\{\min_{v\in L} q_v\}$.
In this paper we consider a well-studied refinement of this approach, known as the
\emph{leximin} solution~\cite[Sec.\ 3.3]{Moulin03}. Since there may be multiple vectors 
$\mathbf{q}$ with the same minimum happiness guarantee, $\min_{v\in L} q_v$, the leximin solution chooses among them one that also maximizes the second smallest happiness value. In case of ties, this process is repeated for the third smallest happiness, and so on, until a unique vector is identified. More generally, if the elements of each feasible vector $\mathbf{q}\in Q$ are sorted in non-decreasing order, then the leximin solution is the vector that is greater than all others from a lexicographic order standpoint.

Given the vector $\mathbf{q}$ of \igf values that correspond to the leximin solution, we can just
run one of the integer linear programs defined above to get the balanced set, but how do we compute
this leximin vector $\mathbf{q}$? A simple way to achieve this is to use binary search in order to first identify the set of maximin solutions. For each value of $q\in [0, 1]$ we can check the feasibility of the integer linear program if we use that same $q$ value for all the groups. Starting from $q=0.5$, if the solution is feasible we update the value to $q=0.75$, whereas if it is infeasible we update it to $q=0.25$, and repeat recursively. Once the largest value of $q$ that can be guaranteed for all groups has been identified, we check which group $I_v$ it is for which this constraint is binding, we fix $q_v = q$, and continue similarly for the remaining groups, aiming to maximize the second smallest \igf value, and so on.

In the experiments of the following section, we compute the leximin solution on real datasets and evaluate its \igf and utility. 
\section{Experimental Evaluation}
\label{sec:experiment}
In this section, we describe the results of an experimental evaluation with three empirical goals: (1) to understand the conditions that lead to reduced \igf; (2) to ascertain the feasibility of mitigating the observed imbalance with the help of \igf constraints; (3) to explore the trade-offs between \igf and utility.

\begin{figure*}[t!]
  \centering
  \includegraphics[width=0.78\linewidth]{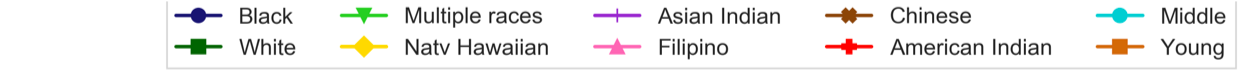}\vspace{-1.2em}
  \subfloat[Diversity constraints only]{
    \includegraphics[width=0.24\linewidth]{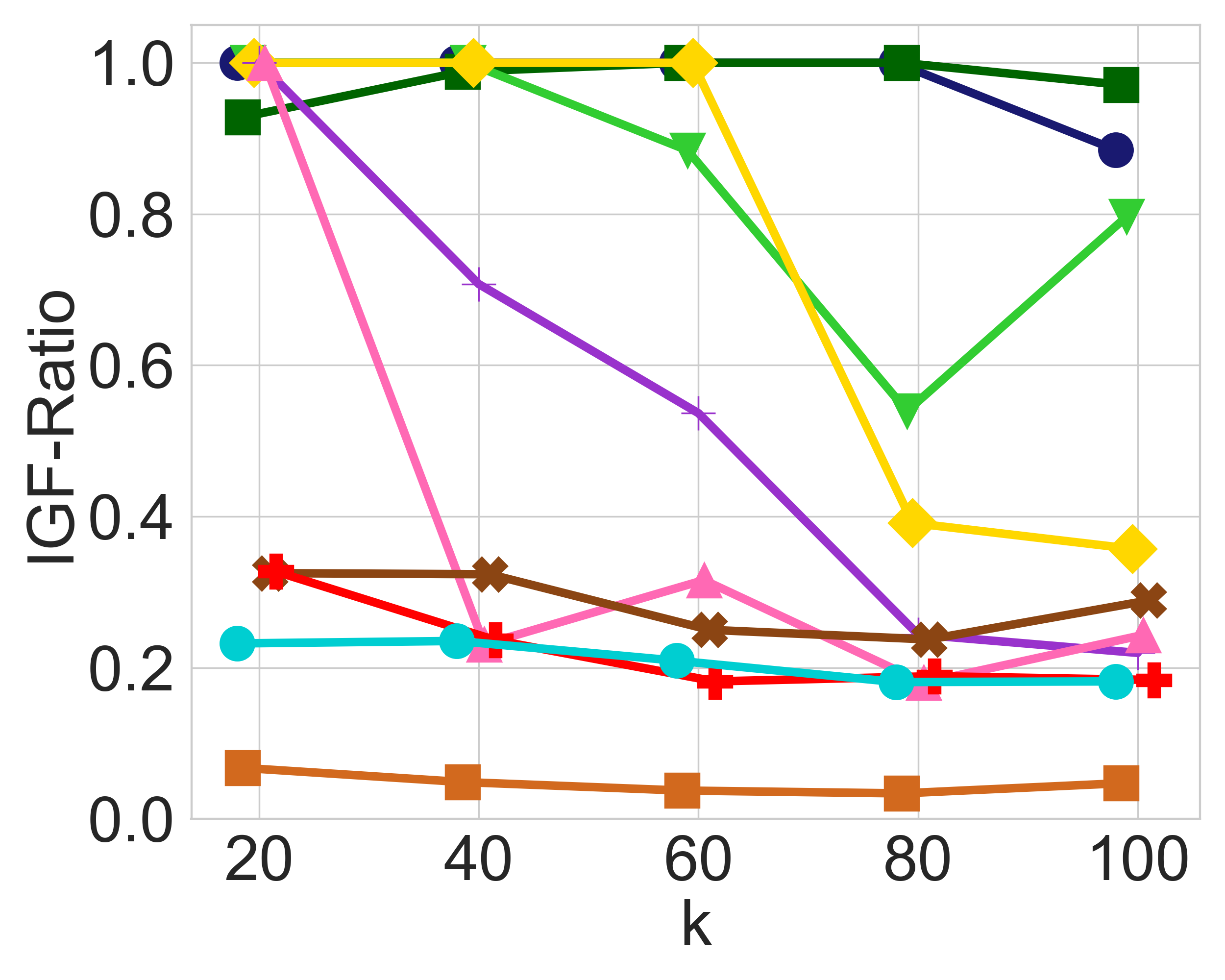}\label{fig:meps_ratio_before}
  }\hfill
  \subfloat[Adding balance constraints]{
    \includegraphics[width=0.24\linewidth]{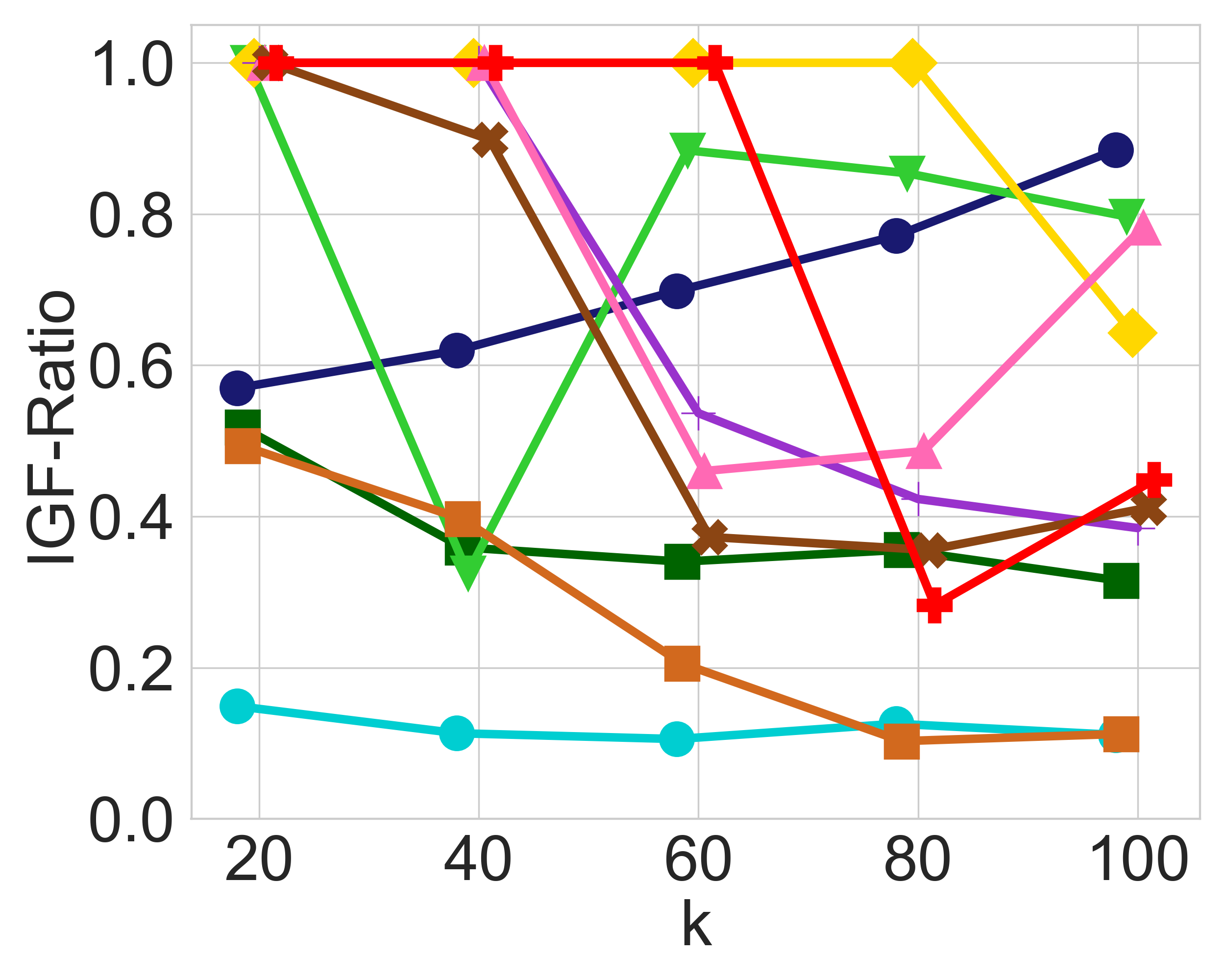}\label{fig:meps_ratio_after}
  }\hfill
  \subfloat[Diversity constraints only]{
      \includegraphics[width=0.24\linewidth]{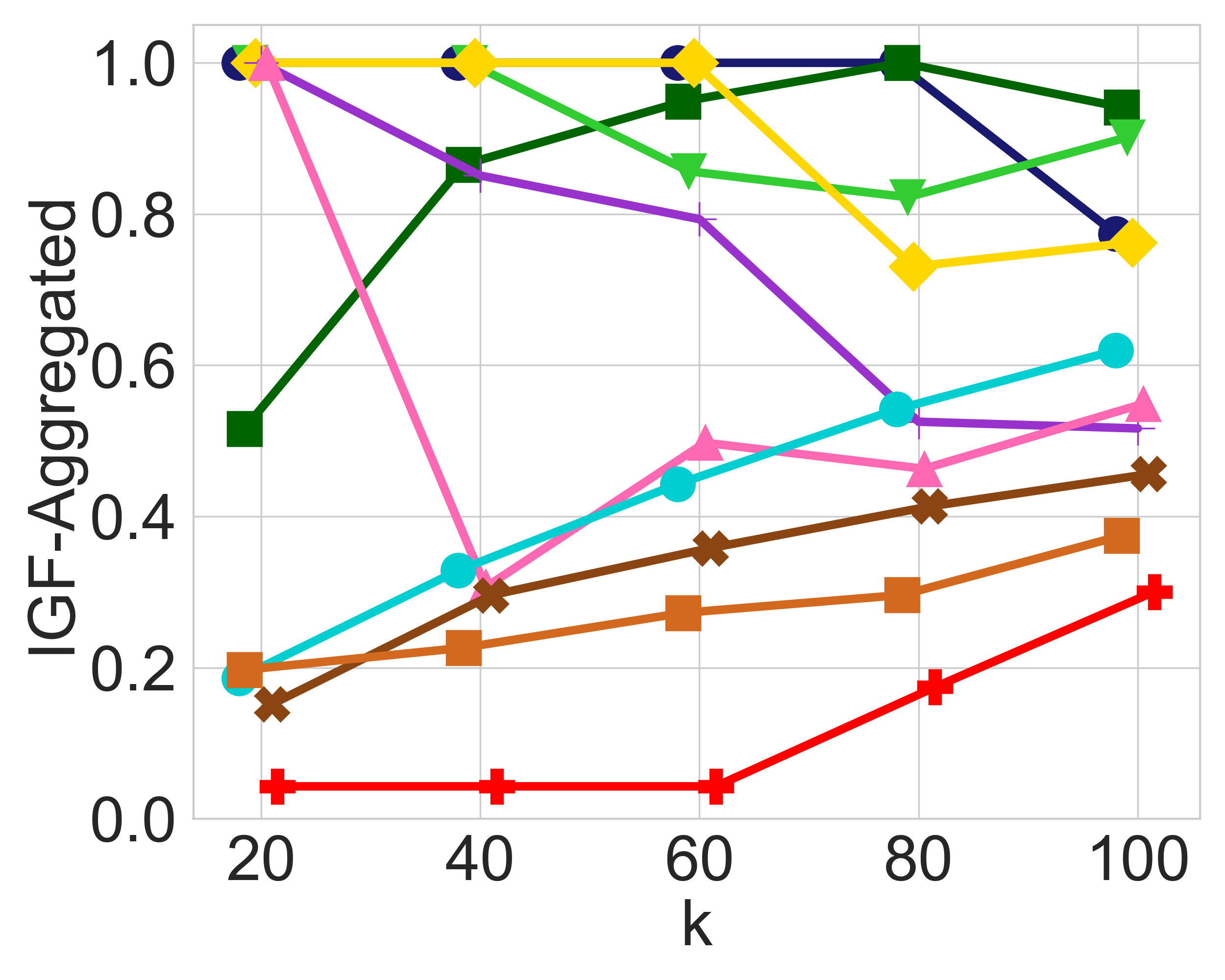}\label{fig:meps_agg_before}
  }\hfill
  \subfloat[Adding balance constraints]{
      \includegraphics[width=0.24\linewidth]{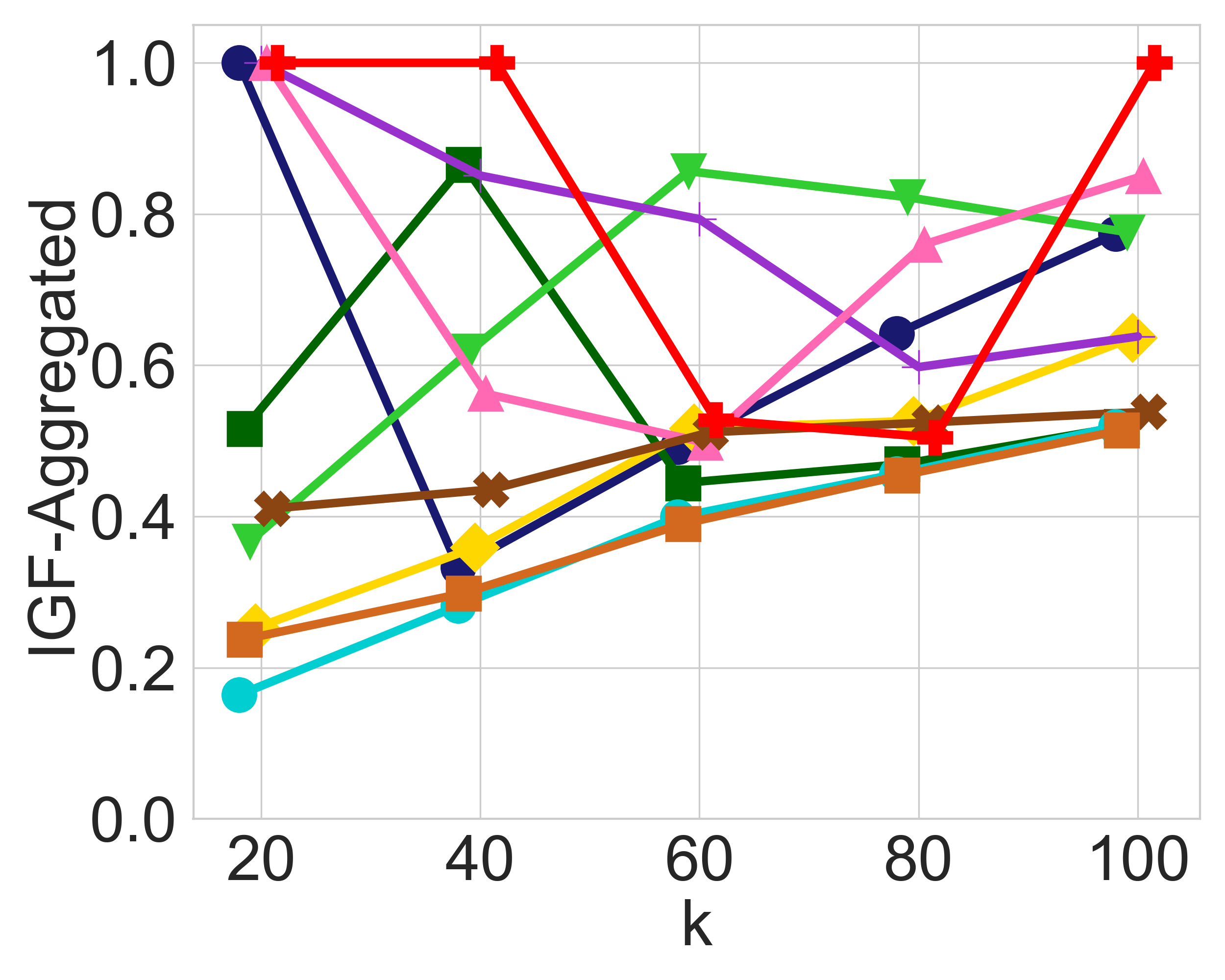}\label{fig:meps_agg_after}
  }
  \caption{In-group fairness for groups defined by {\tt race} and {\tt age} in the MEPS dataset with $5,110$ items, selecting top-$k$ candidates as a ranked list, with diversity constraints on each of the 8 possible {\tt race} groups and 2 {\tt age} groups.}
  \label{fig:meps_ranking}
\end{figure*}
\begin{figure*}[t!]
	\centering
	\includegraphics[width=0.78\linewidth]{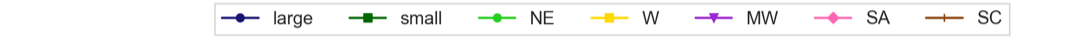}\vspace{-1.2em}
	\subfloat[Diversity constraints only]{
		\includegraphics[width=0.24\linewidth]{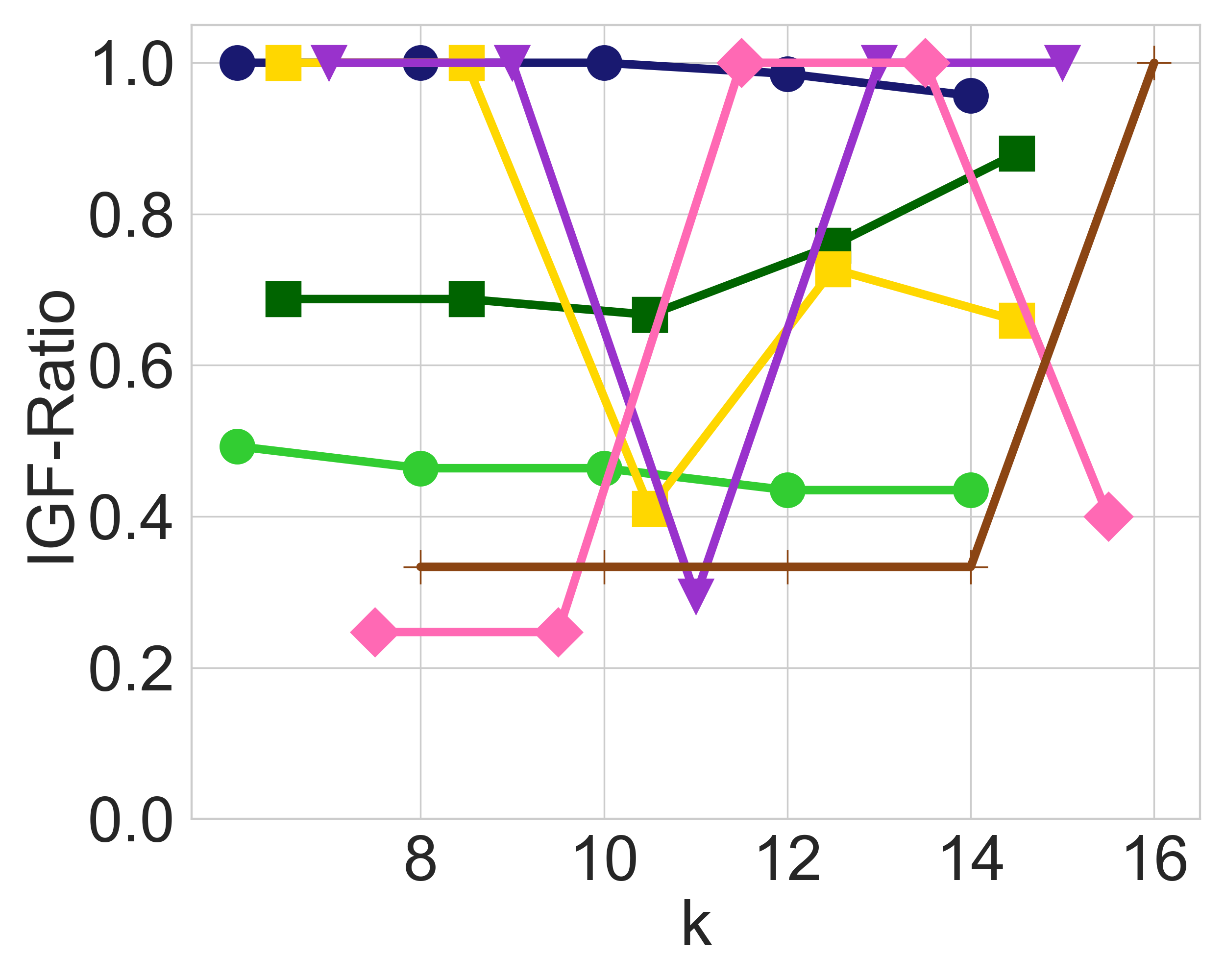}\label{fig:cs_ratio_before}
	}\hfill
	\subfloat[Adding balance constraints]{
		\includegraphics[width=0.24\linewidth]{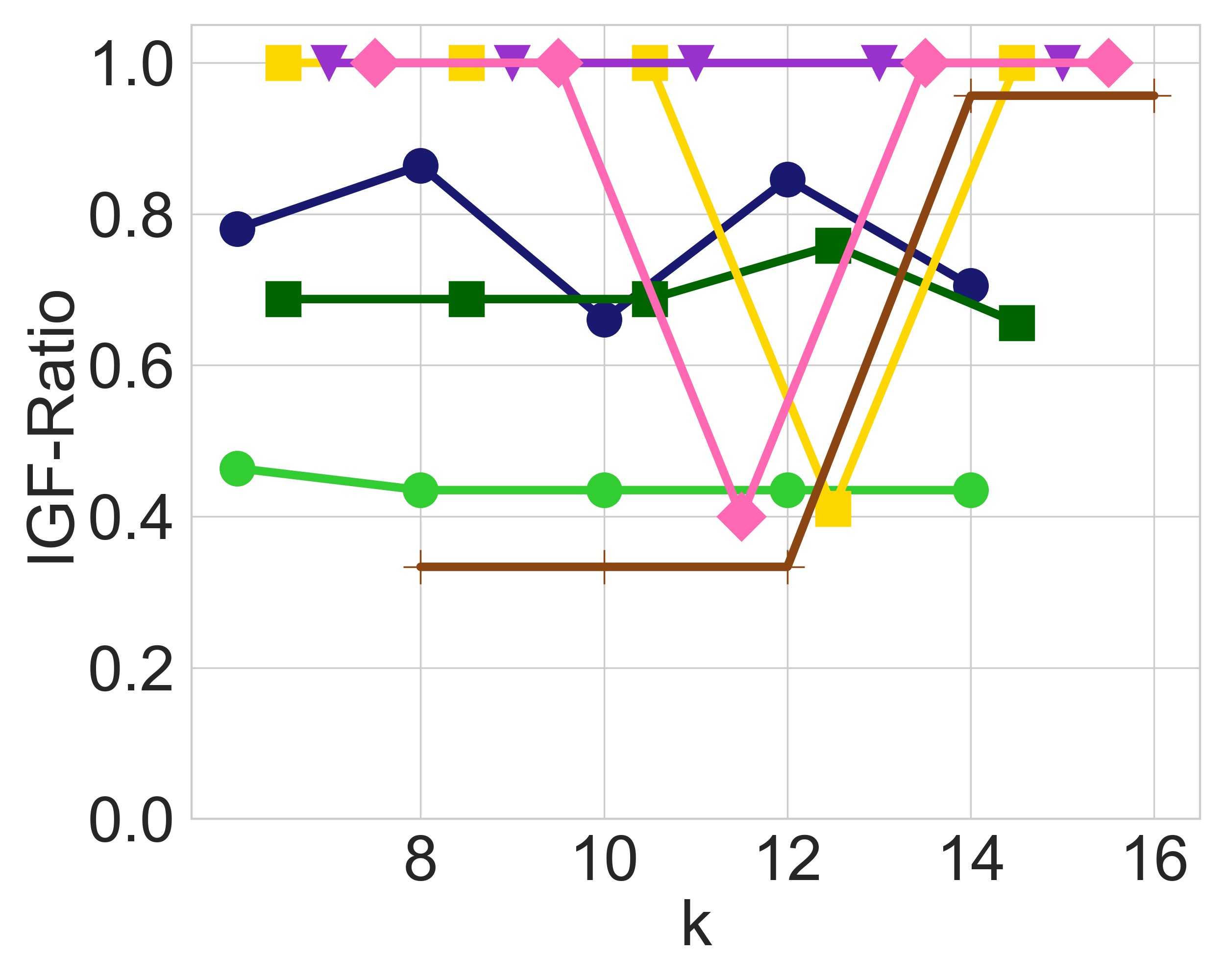}\label{fig:cs_ratio_after}
	}\hfill
	\subfloat[Diversity constraints only]{
		\includegraphics[width=0.24\linewidth]{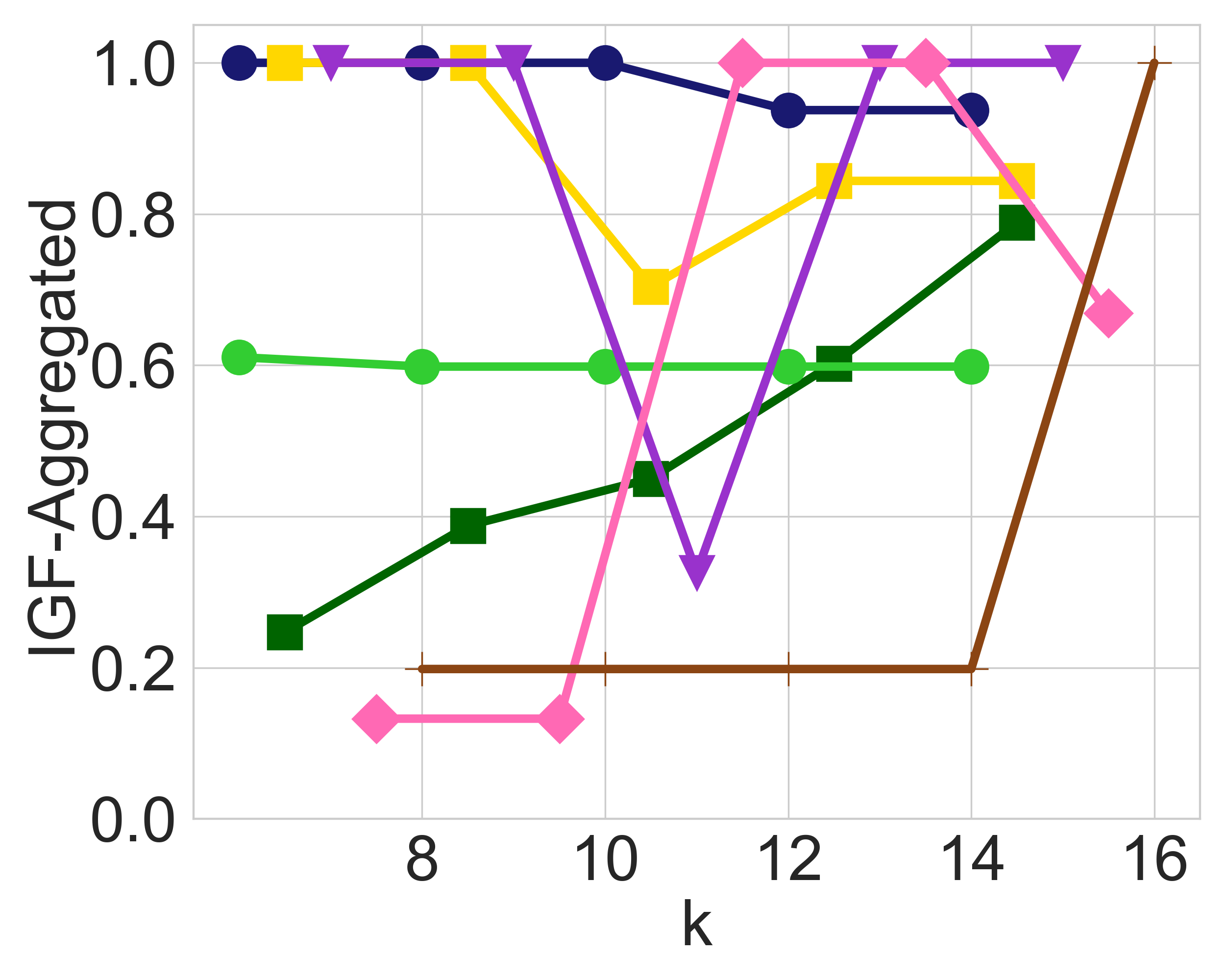}\label{fig:cs_agg_before}
	}\hfill
	\subfloat[Adding balance constraints]{
		\includegraphics[width=0.24\linewidth]{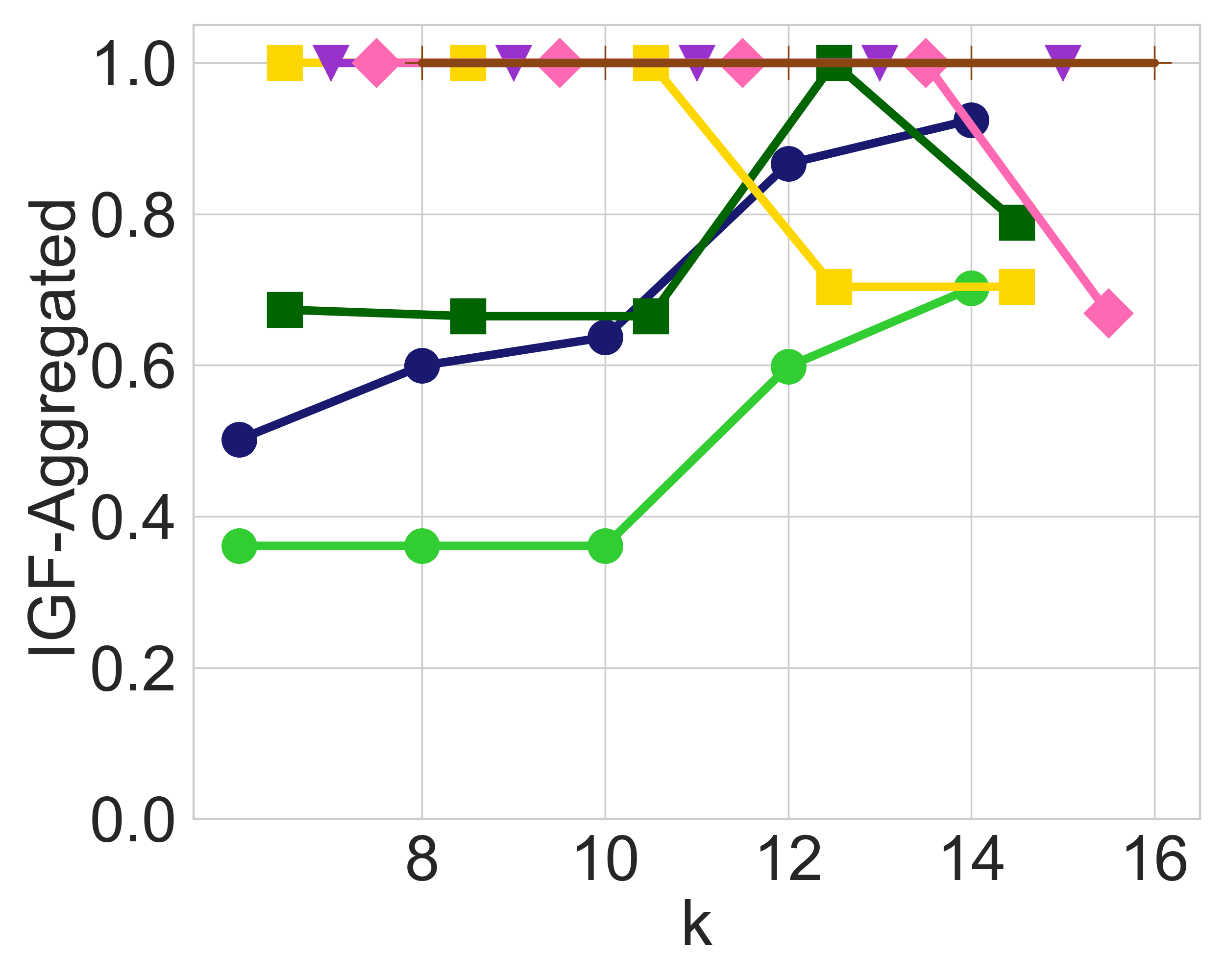}\label{fig:cs_agg_after}
	}
	\caption{In-group fairness for groups defined by {\tt department size} and {\tt area} in the CS dataset with $51$ items, selecting top-$k$ candidates as a ranked list, with diversity constraints on each of the 2 possible {\tt size} groups and 5 {\tt area} groups.}
	\label{fig:cs_ranking}
\end{figure*}
\begin{figure*}[t!]
  \centering
  	\subfloat[Distribution of utilization score in MEPS dataset]{
  	\includegraphics[width=0.48\linewidth]{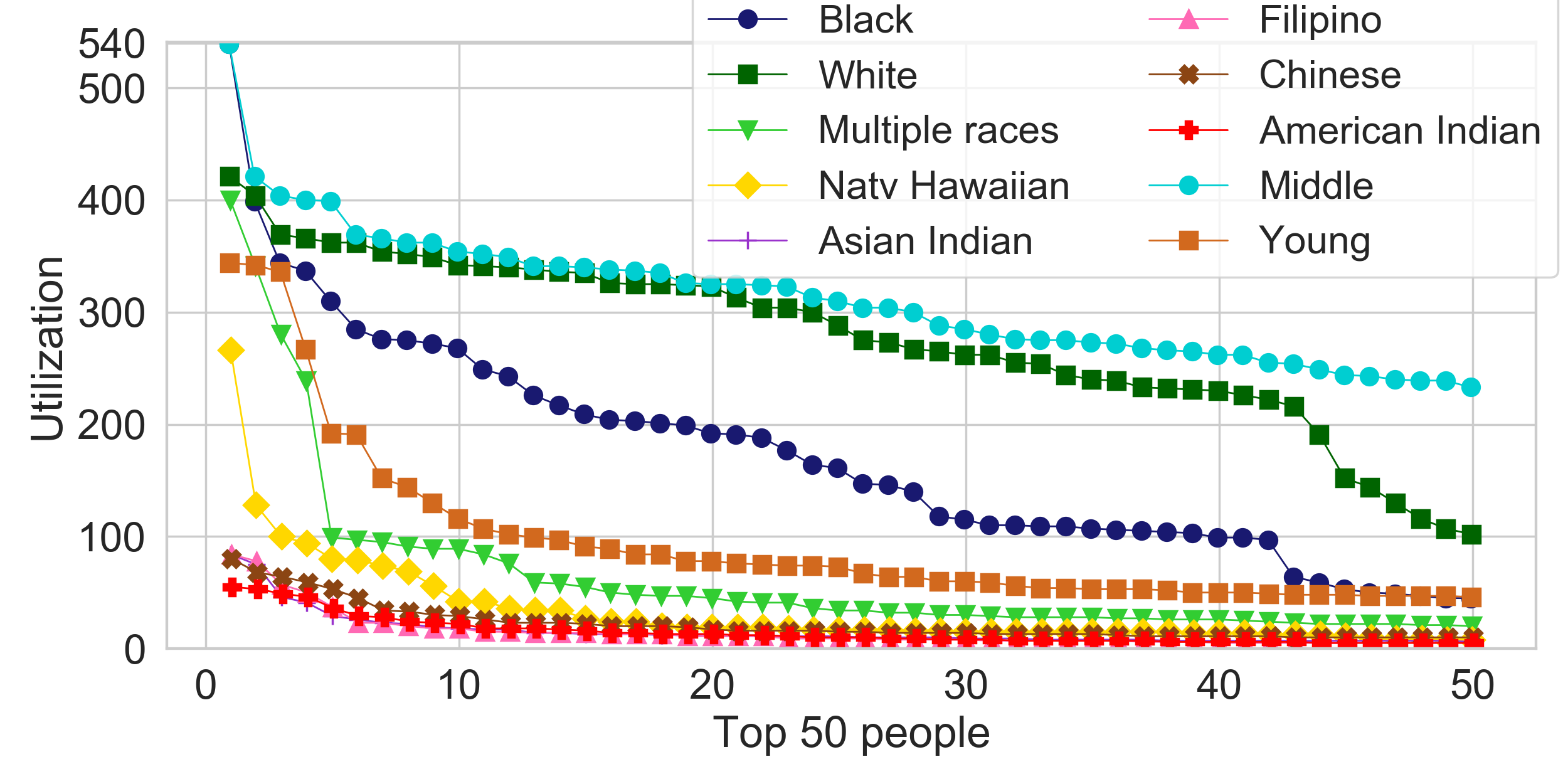}\label{fig:meps_score}
  }\hfill
  \subfloat[Distribution of publication count in CS dataset]{
  	\includegraphics[width=0.48\linewidth]{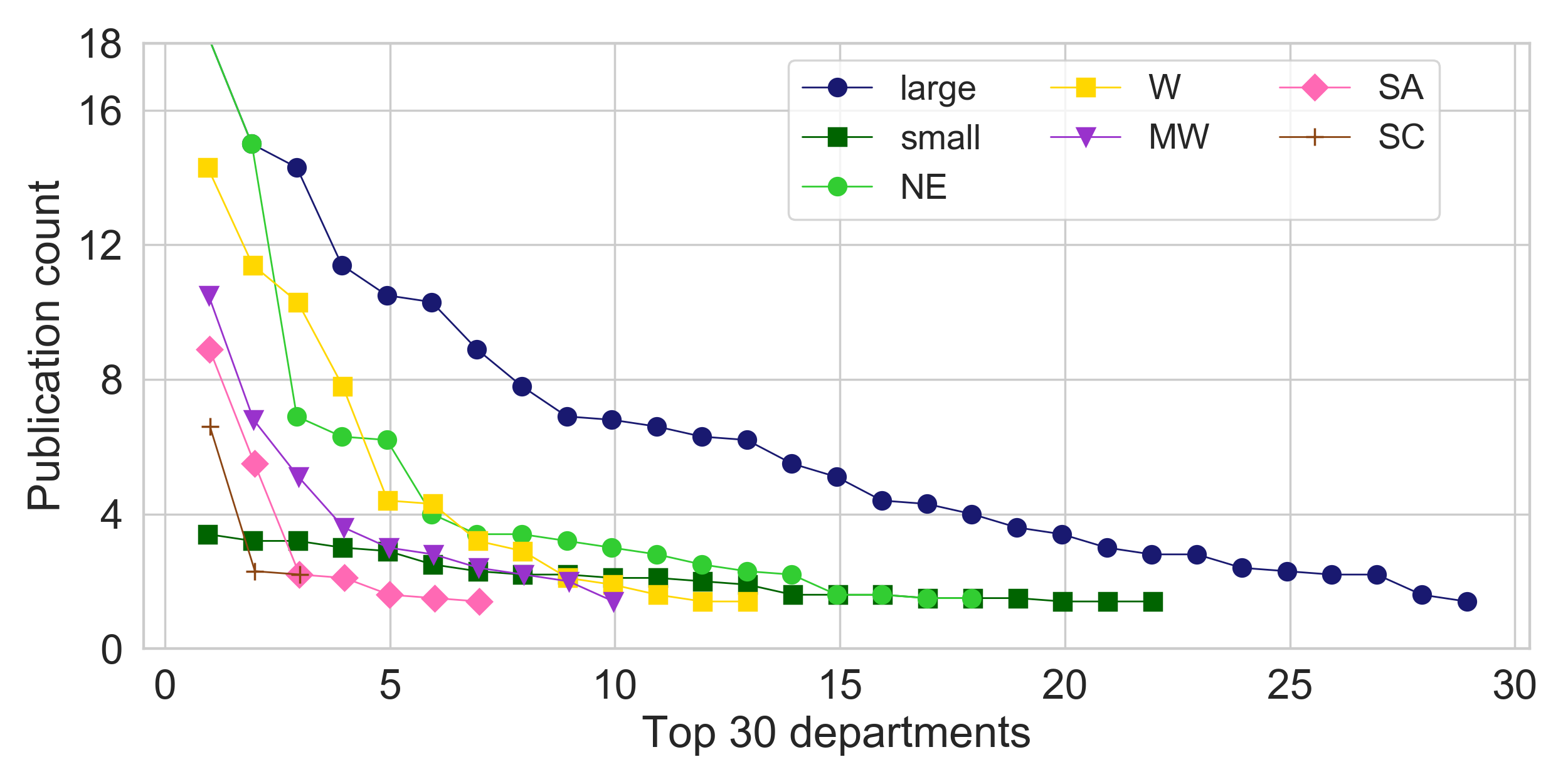}\label{fig:cs_score}
  }
  \caption{Score distribution for groups in MEPS and CS dataset.}
  \label{fig:real_scores}
\end{figure*}
\begin{table*}[t]
	\centering
	\resizebox{0.98\linewidth}{!}{
		\begin{tabular}{||c|c|c|c|c|c|c|c|c|c|c|c|c|c|c|c||}
		    \cline{1-16}
			\multirow{2}{*}{Dataset} & \multicolumn{3}{|c|}{$k_1$} & \multicolumn{3}{|c|}{$k_2$} & \multicolumn{3}{|c|}{$k_3$} & \multicolumn{3}{|c|}{$k_4$} & \multicolumn{3}{|c||}{$k_5$} \\
			\cline{2-16}
			& diversity & ratio & agg & diversity & ratio& agg & diversity & ratio & agg & diversity& ratio & agg & diversity & ratio & agg\\
			\cline{0-15}
			MEPS & 28\% & 5\%  & 3\% & 34\% & 7\% & 3\% & 34\%  & 5\% & 1\% & 33\% & 3\% & 3\% & 28\% & 4\% & 5\% \\
			\cline{0-15}
			CS  & 20\% & 2\%  & 9\% & 18\% & 1\% & 8\% & 17\% & 3\% & 9\% & 12\% & 1\% & 0\% & 11\% & 3\% & 1\% \\
			\cline{0-15}
	\end{tabular}}
	\caption{Percentage of optimal utility lost due to diversity constraints and \igf constraints in MEPS and CS datasets.}
	\label{tab:exp_utility}
\end{table*}

\subsection{Datasets}
\label{sec:dataset}
Our experimental evaluation was conducted on 
the following two real datasets:

\paragraph*{Medical Expenditure Panel Survey (MEPS)} is a comprehensive source of individual and household-level information regarding the amount of health expenditures by individuals from various demographic or socioeconomic groups~\cite{cohen2009medical,coston2019fair}.  MEPS is ideal for our purposes as it includes sensitive attributes such as race and age group for each individual. Each candidate's score corresponds to the {\tt utilization} feature, defined in the IBM AI Fairness 360 toolkit~\cite{bellamy2018ai} as the total number of trips requiring medical care, and computed as the sum of the number of office-based visits, the number of outpatient visits, the number of ER visits, the number of inpatient nights, and the number of home health visits.

High-utilization respondents (with {\tt utilization} $\geq 10$) constitute around 17\% of the dataset.  MEPS  includes  survey data of more than 10,000 individuals. We use data from Panel 20 of calendar year 2016, and select the top 5,110 individuals with {\tt utilization} $> 5$ as our experimental dataset.  
We focus on two categorical attributes: {\tt race} (with values ``White'', ``Black'', ``Multiple races'', ``Native Hawaiian'', ``Asian Indian'', ``Filipino'', ``Chinese'', and ``American Indian'') and {\tt age} (``Middle'' and ``Young''). 

\paragraph*{CS department rankings (CS)} contains information about 51 computer science departments in the US~\cite{csrankings.org}. We use {\tt publication count} as the scoring attribute and select two categorical attributes: {\tt department size} (with values ``large'' and ``small'') and {\tt area} (with values ``North East'', ``West'', ``Middle West'', ``South Center'', and ``South Atlantic''). Unlike the MEPS dataset, CS is a relatively small dataset, but our experiments show that the \constr and the balancing operations exhibit similar behavior. For this dataset we ask for a ranking of the top CS departments, while ensuring that departments of different size and location are well-represented.

\subsection{Experimental Evidence of Imbalance}
\label{sec:evidence}
Using the \igf measures defined in Section~\ref{sec:constr}, we now experimentally evaluate the extent to which \igf is violated in real datasets, as well as how well-balanced it is across the different groups of candidates. Figures~\ref{fig:meps_ratio_before} and \ref{fig:meps_agg_before} exhibit the \ratio and \agg values, respectively, for each of the 8 race groups and 2 age groups of the MEPS dataset, when a ranked list of the top-$k$ candidates is requested, for $k\in \{20, 40, 60, 80, 100\}$. For every prefix of the ranked list, \constr ensure each race and age is represented in that prefix in approximate correspondence with their proportion in the input.  For example, at least $11$ candidates among the top-$20$ are from the ``Middle'' age group, and at least $8$ are from the ``Young'' age group.  

In both Figures~\ref{fig:meps_ratio_before} and \ref{fig:meps_agg_before}, we observe that \constr cause a significant amount of in-group unfairness with respect to the two metrics, leading to values below 0.1 for some groups for both \ratio and \agg. In particular, the young age group and the American Indian race are the two groups with the lowest \igf values in both cases. Apart from low \igf values, we also observe a significant amount of imbalance, since some other groups actually receive very high \igf values for both measures. 

At this point, it is worth providing some intuition regarding why imbalance might arise to begin with. To help us with this intuition we provide Figure~\ref{fig:meps_score}, which exhibits the distribution of the scores for each group in MEPS (both {\tt race} and {\tt age}). From this plot, we can deduce that the individuals in the American Indian group tend to have lower scores compared to other ethnic groups, and that the young age group also has lower scores than the middle age group. However, \constr require that these groups are represented in the output as well. 

An important observation is that selecting a young American Indian in the outcome would satisfy two binding diversity  constraints, while using just one slot in the outcome. This essentially ``kills two birds with one stone'', compared to an alternative solution that would dedicate a separate slot for each minority. The slot that this solution ``saves'', could then be used for a high scoring candidate that is not part of a minority, leading to a higher utility. Therefore without any \igf constraints, the utility-maximizing solution is likely to include a young American Indian who is neither one of the top-scoring candidates in the young group, nor one of the top-scoring American Indians, thus introducing in-group unfairness to both of these groups.

This undesired phenomenon --- the selection of low-quality candidates that satisfy multiple diversity constraints --- is more likely to impact low-scoring groups and may thus disproportionately affect historically disadvantaged minorities, or historically undervalued entities. We observe this in the CS dataset, where small departments, and departments located in South Center and South Atlantic areas, experience higher in-group unfairness before our mitigation (see Figures~\ref{fig:cs_ratio_before},  \ref{fig:cs_agg_before}, and \ref{fig:cs_score}).
 
\subsection{The Impact of Leximin Balancing}
\label{sec:motivation_exp}
Having observed the imbalance that may arise in the output of an unrestricted utility-maximizing algorithm, and having explained how this may systematically adversely impact historically disadvantaged groups, we now proceed to evaluate the impact of the leximin solution. In all the Figures~\ref{fig:meps_ratio_after}, \ref{fig:meps_agg_after}, \ref{fig:cs_ratio_after}, and \ref{fig:cs_agg_after} that are generated by the leximin solution, we see a clear improvement compared to their counterparts, before any balancing was introduced. Recall that the leximin solution's highest priority is to maximize the minimum \igf over all groups, so looking, for instance, at Figure~\ref{fig:meps_agg_after} and comparing it with Figure~\ref{fig:meps_agg_before}, it is easy to see that the minimum \agg value has strictly increased for every value of $k$. Note that this does not mean that every group is better-off. For instance, the White group is significantly worse-off in Figure~\ref{fig:meps_agg_after}, especially for larger values of $k$. However, this drop in the \igf of the White group enabled a very significant increase for the American Indian group, and a noticeable increase for the Chinese and young age groups, which suffered the most in-group unfairness prior to the balancing operation. As a result, the \igf values after the leximin balancing operation are all much more concentrated in the middle of the plot, instead of resembling a bi-modal distribution of high and low \igf values as was the case before, in Figure~\ref{fig:meps_agg_before}.

We observe very similar patterns in all the other applications of balancing as well, exhibiting a consistently better-balanced \igf across groups. Before we conclude, we also show that this significant improvement came without a very significant loss in utility.

\paragraph*{The Price of Balance}
Just like the enforcement of \constr leads to a drop in utility, the same is also true when introducing the \igf constraints 
in order to reach a more balanced distribution. To get a better sense of the magnitude of this loss in utility to achieve
fairness, in Table~\ref{tab:exp_utility} we show the percentage loss caused by \igf and \constr, measured against the optimality 
utility. Note that the $k_1, ..., k_5$ in the columns correspond to $[20, 40, 60, 80, 100]$ for MEPS and to $[8, 10, 12, 14, 16]$ 
for CS. Therefore, the entry in the first columns and first row of this table should be interpreted as follows:
for the MEPS dataset and a ranking request with $k=20$, enforcing \constr leads to a loss of $28\%$ in utility, compared to
the outcome without any such constraints. The entry to its right ($5\%$), is the additional loss in utility caused if, on top
of the \constr, we also enforce the leximin constraints for the \ratio measure. Similarly, the next entry to the right ($3\%$) 
is the additional loss in utility caused if, on top of the \constr, we also enforce the leximin constraints for the \agg measure.
We note that, compared to the $28\%$ of utility loss required by \constr alone, the utility loss due to balancing \igf is actually
quite small, and we observe this to be the case for both datasets and all $k$ values in our experiments.

As a final remark we note that, despite the fact that integer linear programs are not polynomial time solvable in general,
the computational cost involved in computing the utility-maximizing outcome for a given vector of $q_v$ values  
was not too significant, and the standard library solvers were quite efficient. Even for the most time-consuming case, which
was the \agg measure and the MEPS dataset, the solver would complete in a matter of a few minutes. 

\section{Related Work}
\label{sec:related}

Yang and Stoyanovich were the first to consider fairness in ranked outputs~\cite{yang2017measuring}. They focused on a single binary sensitive attribute, such as male or female gender, and minority or majority ethnic group. They then proposed three fairness measures, each quantifying the relative representation of protected group members at discrete points in the ranking (e.g., top-$10$, top-$20$, etc.), and compounding these proportions with a logarithmic discount, in the style of information retrieval. A follow-up work~\cite{DBLP:journals/corr/ZehlikeB0HMB17} developed a statistical test to ascertain whether group fairness holds in a ranking, also with respect to a single binary sensitive attribute, and proposed an algorithm that mitigates the lack of group fairness.  They further proposed the notions of in-group monotonicity and ordering utility that are similar in spirit to our in-group fairness.  The imbalance in terms of in-group fairness does not arise in the set-up of Zehlike et al., where only a single sensitive attribute is considered.

\cite{stoyanovich2018online} considered online set selection under label-representation constraints for a single sensitive attribute, and posed the Diverse $k$-choice Secretary Problem: pick $k$ candidates, arriving in random order, to maximize utility (sum of scores), subject to diversity constraints.  These constraints specify the lowest ($\ell_v$) and highest ($h_v$) number of candidates from group $v$ to be admitted into the final set. The paper developed several set selection strategies and showed that, if a difference in scores is expected between groups, then these groups must be treated separately during processing.  Otherwise, a solution may be derived that meets the constraints, but that selects comparatively lower-scoring members of a disadvantaged group --- it lacks balance.     

Trade-offs between different kinds of accuracy and fairness objectives for determining risk scores are discussed in~\cite{DBLP:conf/innovations/KleinbergMR17}.  The authors use the term {\em balance} to refer to performance of a method with respect to members of a particular class (positive or negative), and note that ``the balance conditions can be viewed as generalizations of the notions that both groups should have equal false negative and false positive rates.''  Our use of the term ``balance'' is consistent with their terminology, as it applies to in-group fairness.

The work of~\cite{DBLP:conf/icalp/CelisSV18} conducts a theoretical investigation of ranking with diversity constraints of the kind we consider here, for the general case of multiple sensitive attributes.  They prove that even the problem of deciding feasibility of these constraints is NP-hard. They provide hardness results, and develop exact and approximation algorithms for the constrained ranking maximization problem, including a linear program and a dynamic programming solution. These algorithms also allow for the scores of the candidates to be position-specific. The novelty of our work compared to that of Celis et al. is that we focus on the imbalance in terms of in-group fairness, develop methods for mitigating the imbalance, and provide an empirical study of these methods.

Diversity, as a crucial aspect of quality of algorithmic outcomes, has been studied extensively in Information Retrieval~\cite{Agrawal:2009:DSR:1498759.1498766,clarke2008novelty} and content recommendation~\cite{kaminskas2017diversity,vargas2011rank}. See also~\cite{drosou2017diversity} for a recent survey of diversity in set selection tasks, including also a conceptual comparison of diversity and fairness, in the sense of statistical parity.

\section{Conclusion and Open Problems}
\label{sec:conc}
In this paper we identified the lack of \igf as an undesired consequence of maximizing total utility subject to diversity constraints, in the context of set selection and ranking. We proposed two measures for evaluating \igf, and developed methods for balancing its loss across  groups. We then conducted a series of experiments to better understand this issue and the extent to which our methods can mitigate it. This paper opens up many interesting research directions, both empirical and theoretical.

From an empirical standpoint, it would be important to develop a deeper understanding of the aspects that may cause disproportionate in-group unfairness. In our experimental evaluation we observed that, all else being equal, a larger difference in expected scores between disjoint groups leads to a higher imbalance in terms of in-group fairness. In the future, we would like to identify additional such patterns that may lead to systematic imbalance, especially when this may disproportionately impact disadvantaged groups. 

From a theoretical standpoint, it would be interesting to understand the extent to which polynomial time algorithms can approximate the optimal utility or approximately satisfy the combination of diversity and \igf constraints. In this paper we restricted our attention to finding exact solutions to the defined optimization problems and, since even simplified versions of these problems are NP-hard, we had to resort to algorithms without any appealing worst-case guarantees. In~\cite{DBLP:conf/icalp/CelisSV18}, the authors considered the design of approximation algorithms for closely related problems involving diversity constraints, but without any \igf constraints. On the other hand, \cite{DBLP:conf/icalp/CelisSV18} also consider ranking problems where the candidates' scores can be position-dependent. Extending our framework to capture these generalizations is another direction for future work.

Finally, moving beyond the leximin solution, one could consider alternative ways to choose the \igf vector $\mathbf{q}$, such as maximizing the Nash social welfare, an approach that has recently received a lot of attention~(e.g.,~\cite{CG18} and \cite{CKMP016}).

\paragraph{Acknowledgements.} 
This work was supported in part by NSF Grants No. 1926250, 1916647, and 1755955.
\balance

\newpage
\bibliographystyle{named}
\bibliography{balance}

\end{document}